\documentclass[10pt,twocolumn,letterpaper]{article}

\usepackage{iccv}              %

\definecolor{iccvblue}{rgb}{0.21,0.49,0.74}
\usepackage[pagebackref,breaklinks,colorlinks,allcolors=iccvblue]{hyperref}

\def\paperID{6850} %
\def\confName{ICCV}
\def\confYear{2025}

\title{ToolVQA: A Dataset for Multi-step Reasoning VQA with External Tools}

\author{Shaofeng Yin\quad Ting Lei\quad Yang Liu\thanks{Corresponding author}\\
Wangxuan Institute of Computer Technology, Peking University\\
{\tt\small yin\_shaofeng@stu.pku.edu.cn}\quad {\tt\small \{ting\_lei, yangliu\}@pku.edu.cn}
}

\usepackage[accsupp]{axessibility}
\usepackage{pifont}
\usepackage{graphicx}
\usepackage{multirow}
\usepackage{booktabs}
\usepackage{amsmath}
\usepackage{makecell}
\usepackage{colortbl}
\usepackage{xcolor}
\usepackage{caption} %
\newcommand{\cmark}{\textcolor{green}{\ding{51}}} %
\newcommand{\xmark}{\textcolor{red}{\ding{55}}} %
\usepackage{algorithm}
\usepackage{algpseudocode}
\usepackage{amsfonts}
\usepackage{amssymb}
\usepackage{array}
\usepackage{mflogo}
\usepackage{subcaption}
\usepackage{listings}
\begin{document}

\newcommand{\ting}[1]{\textcolor{magenta}{[Ting: #1]}}
\newcommand{\yang}[1]{\textcolor{blue}{[Yang: #1]}}
\newcommand{\shaofeng}[1]{\textcolor{brown}{[Shaofeng: #1]}}

\maketitle
\begin{abstract}
Integrating external tools into Large Foundation Models (LFMs) has emerged as a promising approach to enhance their problem-solving capabilities. While existing studies have demonstrated strong performance in tool-augmented Visual Question Answering (VQA), recent benchmarks reveal significant gaps in real-world tool-use proficiency, particularly in functionally diverse multimodal settings requiring multi-step reasoning.  
In this work, we introduce ToolVQA, a large-scale multimodal dataset comprising 23K samples, designed to bridge this gap. Unlike previous datasets that rely on synthetic scenarios and simplified queries, ToolVQA features real-world visual contexts and challenging implicit multi-step reasoning tasks, better aligning with real user interactions.
To construct this dataset, we propose ToolEngine, a novel data generation pipeline that employs image-guided Depth-First Search (DFS) with a Longest Common Subsequence (LCS)-based example matching mechanism to simulate human-like tool-use reasoning. 
ToolVQA encompasses 10 multimodal tools across 7 diverse domains, with an average inference length of 2.78 reasoning steps per sample.  
The LLaVA-7B model fine-tuned on ToolVQA not only achieves impressive performance on the ToolVQA test set, but also surpasses the large closed-source model GPT-3.5-turbo on five out-of-distribution (OOD) datasets, showing strong generalizability in real-world tool-use scenarios.
Code is available at https://github.com/Fugtemypt123/ToolVQA-release.

\end{abstract}
    
\section{Introduction}
\label{sec:intro}

Integrating external tools into Large Foundation Models (LFMs)~\cite{qin2024toollearningfoundationmodels} is a promising way to build versatile AI assistants. By leveraging external tools, LFMs can break down complex tasks into smaller, more manageable subtasks, each handled by a specialized tool. Prior works~\cite{schick2023toolformerlanguagemodelsteach, suris2023vipergpt, gao2023assistgptgeneralmultimodalassistant, hu2024visualprogramdistillationdistilling, hu2023avisautonomousvisualinformation} show strong performance on traditional Visual Question Answering (VQA) benchmarks using external tools. 
Recent works~\cite{wang2024gtabenchmarkgeneraltool, ma2024mms, mialon2023gaiabenchmarkgeneralai} highlight the limitations of current LFMs in their tool-use capabilities within challenging contexts. These contexts require functionally diverse multimodal tools and multi-step reasoning processes associated with tool usage, posing major challenges to LFMs’ tool-use proficiency.

\begin{figure}[t]
    \centering
    \includegraphics[width=1.0\linewidth]{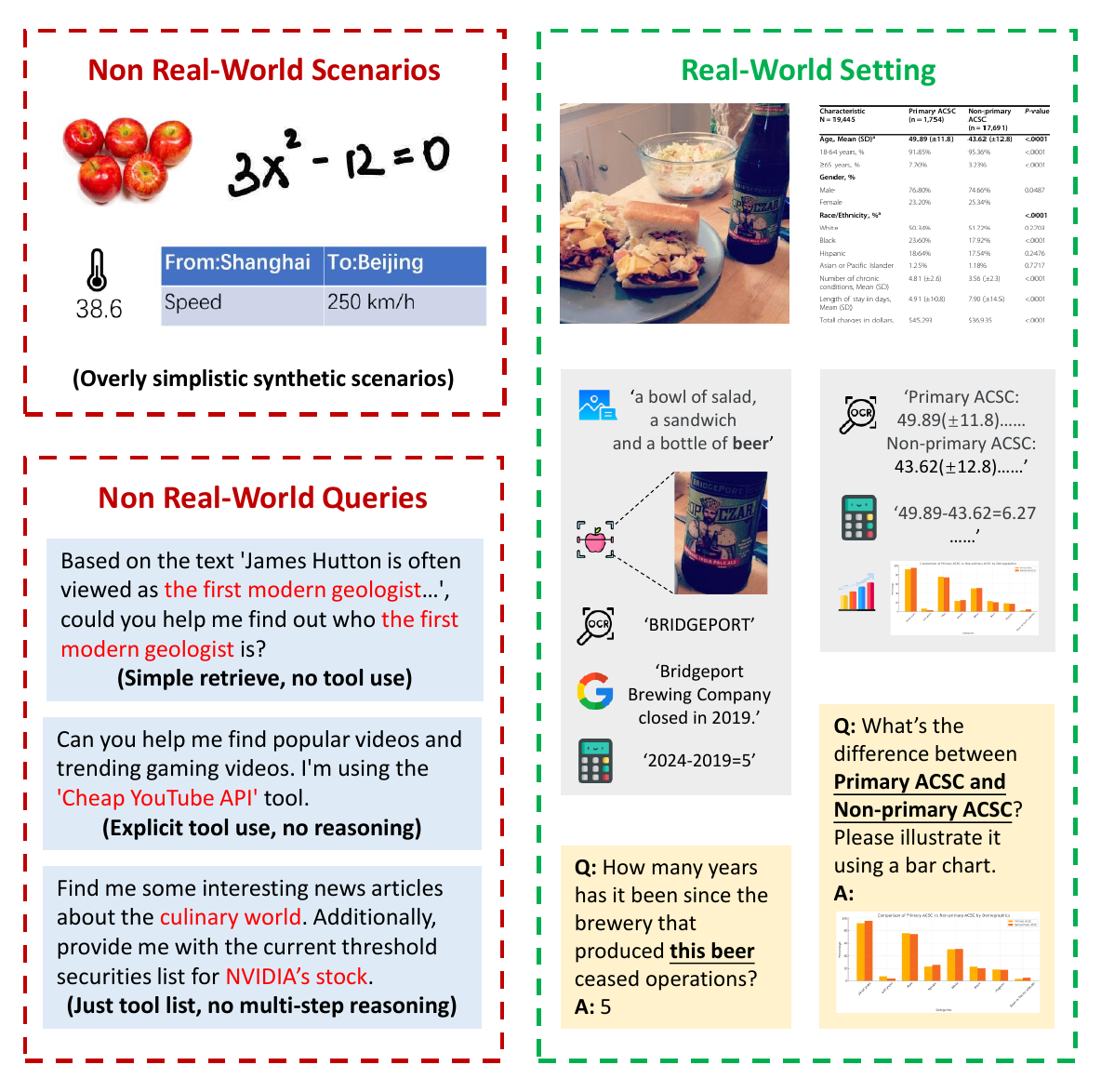}
    \caption{Our real-world setting includes (1) complex visual scenarios with real-world context; (2) challenging queries with an implicit multi-step reasoning process. Existing datasets (left) do not meet these requirements, while our ToolVQA (right) does.}
    \label{fig:teaser}
\end{figure}

One effective approach to enhancing LFM's tool-use capability is fine-tuning on large-scale datasets~\cite{schick2023toolformerlanguagemodelsteach, huang2024metatoolbenchmarklargelanguage, hu2024visualprogramdistillationdistilling, gao2025multimodalagenttuningbuilding}. However, there remains a significant gap between existing large-scale datasets and real-world user needs. Through the study of real user logs, we find that real-world tool usage involves (1) real-world scenarios: complex visual scenarios with real-world context (\eg real-taken photos rather than synthetic images), and (2) real-world queries: challenging queries with an implicit multi-step reasoning process. However, existing datasets often lack these two critical properties, as shown in~\cref{fig:teaser}. 
The scenarios are synthesized instead of real-taken~\cite{wang2024gtabenchmarkgeneraltool,gao2025multimodalagenttuningbuilding}, resulting in simplified scenes that differ from real-world scenarios in complexity.
The queries require only single-step reasoning or explicitly provide hints about the reasoning process~\cite{hu2024visualprogramdistillationdistilling, qin2023toolllm}, such as ``using the Cheap YouTube API tool''.
These setups simplify the task and may lead to a gap
between synthetic and real-world tool usage, which limits our ability to analyze LFMs' performance on this task.  
Additionally, some datasets~\cite{wang2024gtabenchmarkgeneraltool,mialon2023gaiabenchmarkgeneralai} rely on costly human annotations, making them difficult to scale, which limits fine-tuning to enhance LFMs' tool-using capabilities.

To bridge these gaps, we introduce ToolEngine, a data construction pipeline designed to generate multi-step reasoning VQA data for tool usage. ToolEngine employs an image-guided Depth-First Search (DFS) approach~\cite{qin2023toolllm} to simulate the human-like multi-step reasoning process under the guidance of real user context examples. To facilitate multi-step reasoning in data, we introduce a Longest Common Subsequence (LCS)-based example matching mechanism. 
During each step in the iterative search process, this mechanism matches different real-world tool-use examples based on the previous search trajectory, guiding the controller in making the next selection. 
This mechanism enables: (1) Extracting rich visual information from the input image; (2) Integrating this information during the reasoning process to propose challenging queries. By freeing query generation from the constraints of fixed templates~\cite{qin2023toolllm,ma2024mms, gao2025multimodalagenttuningbuilding}, ToolEngine enhances the utilization of visual information, consequently increasing the reasoning complexity of the generated queries.

We leverage ToolEngine to construct ToolVQA, a large-scale multimodal tool-usage dataset comprising 23K samples. We assess data quality via a manual evaluation conducted on a randomly sampled 4k subset of the training set, which yields an accuracy of 90.8\%.
The test set is carefully filtered and fully human-annotated.
As shown in~\cref{tab:comparison}, ToolVQA outperforms prior tool-use datasets across multiple perspectives, providing a more comprehensive representation of real-world scenarios. We evaluate state-of-the-art (SOTA) LFMs on ToolVQA and observe that:
(1) Larger LFMs demonstrate superior reasoning capabilities.
(2) LFMs consistently underperform relative to humans, primarily due to their inability to integrate new information introduced in multi-turn dialogues.
We fine-tune LLaVA-7B~\cite{liu2023visualinstructiontuning} on ToolVQA, and the model demonstrated performance surpassing or approaching that of the large closed-source model GPT-3.5-Turbo on both our test set and five out-of-distribution (OOD) benchmarks—TextVQA~\cite{singh2019vqamodelsread}, TallyQA~\cite{acharya2018tallyqaansweringcomplexcounting}, Infoseek~\cite{chen2023pretrainedvisionlanguagemodels}, TEMPLAMA~\cite{Dhingra_2022}, and GTA~\cite{wang2024gtabenchmarkgeneraltool}, which include unseen tasks and tools. 
In summary, our contributions are as follows:

\begin{enumerate}
    \item \textbf{ToolEngine:} We introduce ToolEngine, a VQA data construction engine that generates reasonable multi-step tool-use trajectories from unannotated images, and constructs challenging question-answer pairs accordingly.
    \item \textbf{ToolVQA:} We propose ToolVQA, a large-scale multimodal tool-use dataset that encompasses 10 multimodal tools across 7 diverse domains, providing a benchmark for evaluating and advancing LFMs' tool-use capabilities on real-world tasks. 
    \item \textbf{Fine-tuning:} We fine-tune LLaVA-7B on ToolVQA to obtain a new tool-use agent, which outperforms large closed-source models like GPT-3.5-Turbo on six benchmarks (ToolVQA and the other five OOD benchmarks).
\end{enumerate}

\begin{table*}[h]
\centering
\begin{tabular}{lcccccccccc}
\toprule
 Dataset & {\makecell[c]{Number \\ of Samples}}& {\makecell[c]{Number \\ of Tools}}& {\makecell[c]{Mutlimodal \\ Inputs?}} & {\makecell[c]{Real-World\\ S/Q Setting?}} & {\makecell[c]{Evaluable\\Answers?}}   & {\makecell[c]{Reasoning\\Complexity$\uparrow$}}  &  \\
\midrule
 ToolBench~\cite{qin2023toolllm} & 126.5K & 3451 & \xmark & - / \xmark & \xmark   & 1.43   \\
 ToolQA~\cite{zhuang2023toolqadatasetllmquestion} & 1.5K & 13 & \xmark & - / \cmark & \cmark   & 1.35    \\
\midrule
 Infoseek~\cite{chen2023pretrainedvisionlanguagemodels} & 1.3M & 1 & \cmark & \cmark / \cmark & \cmark   & 1.13    \\
VPD~\cite{hu2024visualprogramdistillationdistilling}$^*$ & 310.5K & 4 & \cmark & \cmark / \cmark & -  & - \\
M\&m's~\cite{ma2024mms}  & 1.6K & 33 & \cmark & \cmark / \xmark & \xmark   & 1.63    \\
 GAIA~\cite{mialon2023gaiabenchmarkgeneralai}$^\dag$ & 0.5K   & - & \cmark & \cmark / \cmark & \cmark & \textbf{2.61} \\
GTA~\cite{wang2024gtabenchmarkgeneraltool} & 0.2K & 14 & \cmark & \xmark / \cmark & \cmark  & 1.86 \\
MM-Traj~\cite{gao2025multimodalagenttuningbuilding}$^\ddag$ & 20K & 9 & \cmark & \xmark / \cmark & \xmark & 1.77 \\
\midrule
\textbf{ToolVQA (Ours)} & 23.7K & 10 & \cmark & \cmark / \cmark & \cmark   &  \underline{2.38}     \\
\bottomrule
\end{tabular}
\caption{Comparison between ToolVQA and other datasets. \textbf{Real-World S/Q Setting}: whether each sample in the dataset involves: (1) Scenario—utilizes real rather than synthetic multimodal inputs; (2) Query—implicitly embeds tool usage and reasoning within the query.
\textbf{Reasoning Complexity}: the average reasoning depth of humans for each task. 
VPD$^*$ constructs trajectories from existing QA pairs and is not publicly accessible. 
GAIA$^\dag$'s each sample corresponds to a distinct set of human-annotated tools. $^\ddag$MM-Traj generates over-simplified PDF files as multimodal contexts, and their answers are long-form texts that have not been verified for correctness.} 
\label{tab:comparison}
\end{table*}

\section{Related Work}
\label{sec:related}

\subsection{Datasets and Benchmarks for Tool Agent}

With the rise of real-world tool-use applications, many works have attempted to build datasets and benchmarks to evaluate the tool-use capabilities of LFMs from different perspectives. Text-based datasets~\cite{qin2023toolllm, Peng2021RevisitingBA, Li2023APIBankAC, zhuang2023toolqadatasetllmquestion, ruan2024identifyingriskslmagents, huang2024metatoolbenchmarklargelanguage, yang2018hotpotqadatasetdiverseexplainable, yao2023webshopscalablerealworldweb} collect a large number of real-world APIs to build large-scale, diverse text-based tool-use examples. 
On the other hand, multimodal datasets~\cite{ma2024mms, mialon2023gaiabenchmarkgeneralai, wang2024gtabenchmarkgeneraltool, shen2024taskbenchbenchmarkinglargelanguage, acharya2018tallyqaansweringcomplexcounting, singh2019vqamodelsread, chen2023pretrainedvisionlanguagemodels, mani2022pointaskincorporatingpointing, hudson2019gqa, wang2024videocotvideochainofthoughtdataset} integrate visual input into the query process, making queries challenging and more adaptable to real-world needs. However, they either fail to capture real-world scenarios or rely on expensive human annotation. MM-Traj~\cite{gao2025multimodalagenttuningbuilding} is a recent work which also constructs large-scale training data using an automated pipeline, but has the following limitations: (1) Synthesizes over-simplified PDF files as the multimodal contexts; 
(2) Their answers are long-form texts that have not been verified for correctness.

\subsection{LFM-based Tool Agents}
Most recent LFMs have demonstrated impressive reasoning capabilities on numerous zero-shot tasks, which makes it possible to build LFM-based tool agents for real-world tasks. Mainstream approaches include in-context learning~\cite{suris2023vipergpt, castrejon2024hammrhierarchicalmultimodalreact, lu2023chameleon, 
hao2024toolkengptaugmentingfrozenlanguage, wu2024avataroptimizingllmagents, yang2024doraemongptunderstandingdynamicscenes, yang2023mmreactpromptingchatgptmultimodal} and instruction fine-tuning~\cite{schick2023toolformerlanguagemodelsteach, 
qin2023toolllm,
hu2024visualprogramdistillationdistilling, 
patil2023gorilla, 
wang2024mllmtool}.
However, these agents still face challenges in real-world applications. %
Additionally, recent advancements in web navigation agents~\cite{iong-etal-2024-openwebagent,xie2024osworldbenchmarkingmultimodalagents,ma2024laserllmagentstatespace,huq2025cowpilotframeworkautonomoushumanagent,wang2024agentworkflowmemory,zhou2024webarenarealisticwebenvironment,koh2024visualwebarenaevaluatingmultimodalagents} have primarily focused on GUI-based interactions, often designed as browser plugins, but their applicability is limited to web-based tasks. In contrast, our ToolEngine extends to a broader range of real-world scenarios. %

\subsection{Other VQA Datasets}
Visual Question Answering (VQA) is one of the most fundamental tasks in multimodal learning. Early VQA datasets~\cite{agrawal2016vqavisualquestionanswering, zhu2016visual7wgroundedquestionanswering,  lin2014microsoftcoco, 
deng2009imagenet, 
krishna2017visual-genome, 
hudson2019gqa, 
schwenk2022aokvqa} 
primarily focus on simple commonsense question-answering, which typically consists of tasks that humans can easily solve. %
Recently, some studies~\cite{marino2019okvqa,  chen2023pretrainedvisionlanguagemodels, singh2019vqamodelsread, acharya2018tallyqaansweringcomplexcounting, hu2024visualprogramdistillationdistilling} 
begin exploring more challenging forms of VQA, which require various tools like Wikipedia, OCR, Object Detection, etc. However, these datasets typically only evaluate LFM's ability to use one specific tool instead of collaboration between multiple tools, limiting LFM’s multi-step reasoning ability within a tool-agent framework.
In contrast, ToolVQA contains high-quality and diverse multi-step tool usage traces. It encompasses 10 multimodal tools across 7 diverse domains, with an average inference length of 2.78 reasoning steps per sample.

\section{Data Collection}
\label{sec:DataConstruction}

\subsection{Task Formulation}
\label{sec:formulation}

\begin{figure}[h]
    \centering
    \includegraphics[width=0.8\linewidth]{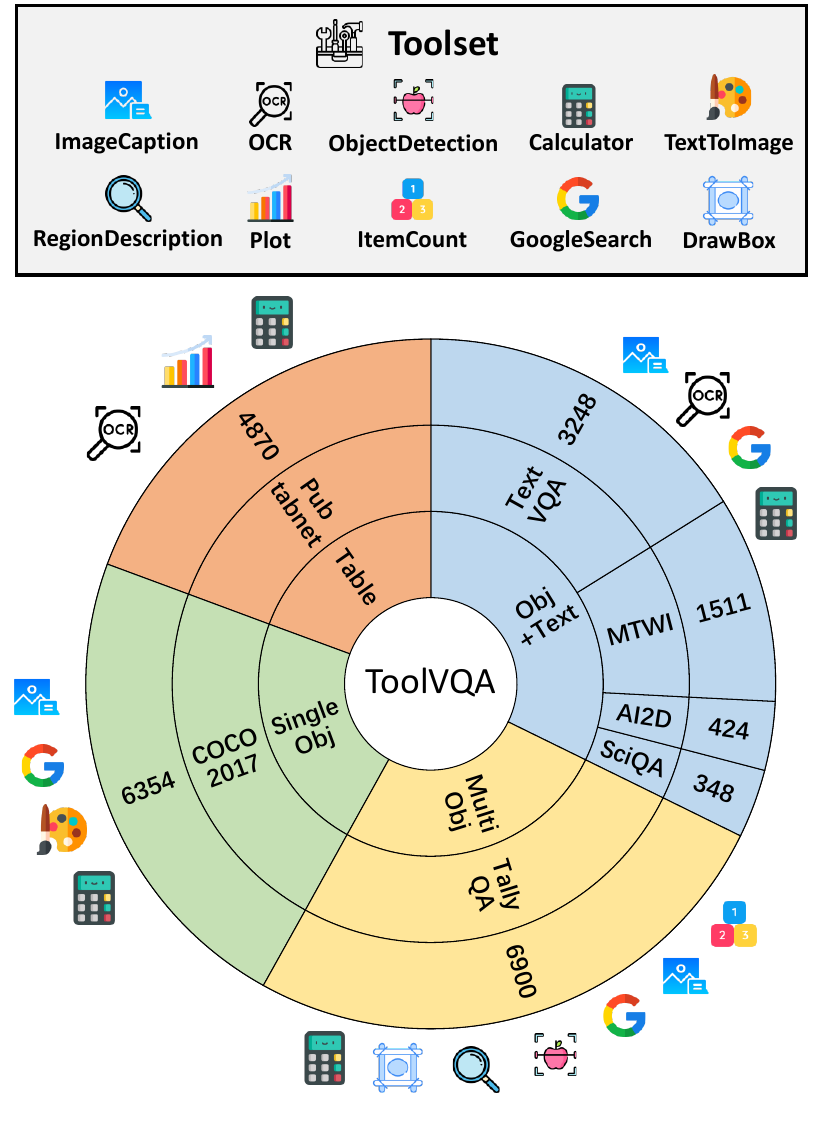}
    \caption{Image sources and corresponding tools of ToolVQA. We filter out overly simplistic tables and images from the data sources.}
    \label{fig:source}
\end{figure}

\begin{figure*}[t]
  \centering
   \includegraphics[width=0.95\linewidth]{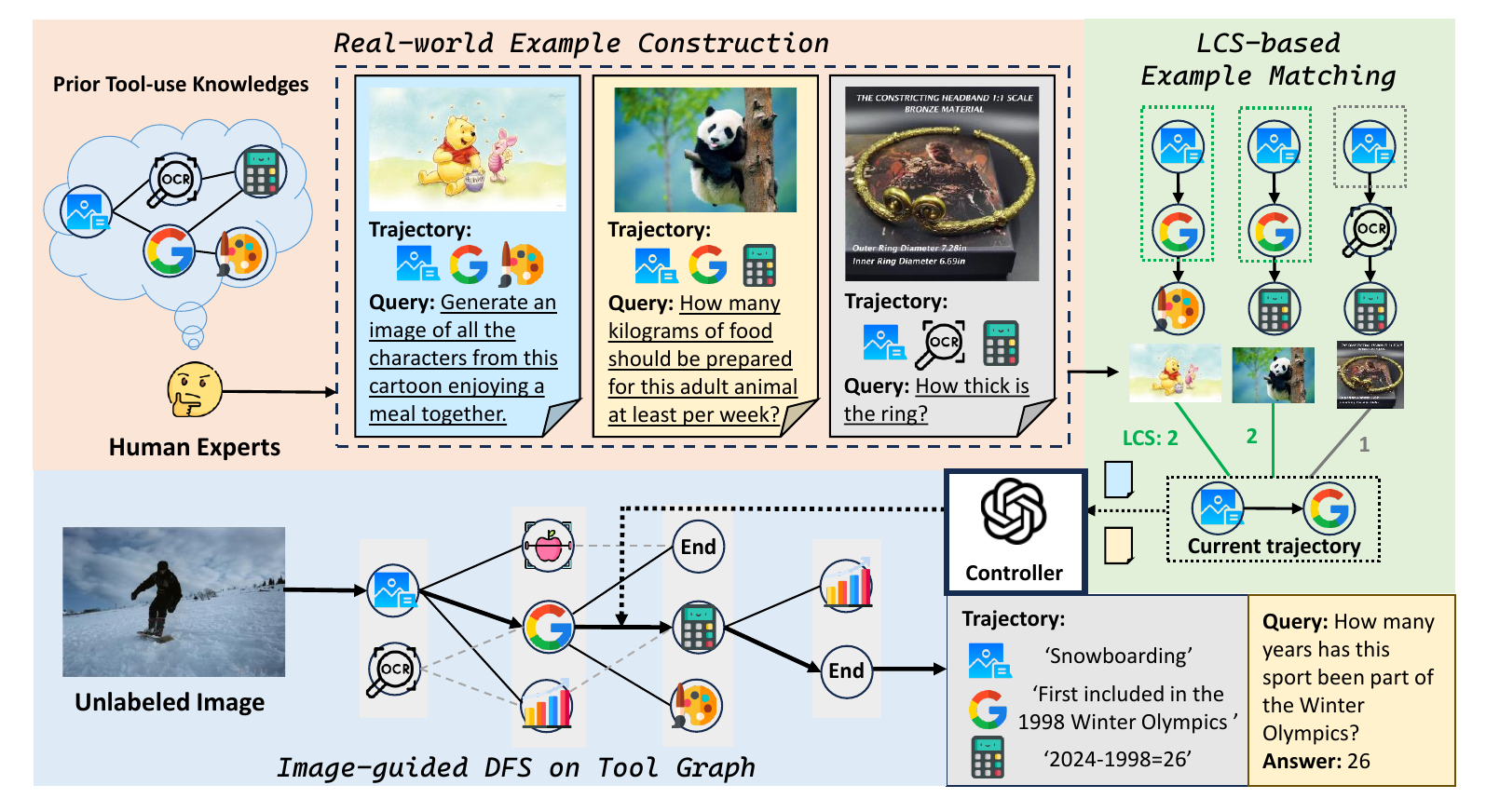}
   \caption{The pipeline of ToolEngine, contains three core components: Real-world Example Construction, Image-guided DFS on Tool Graph, and LCS-based Example Matching. Given an input image, we perform DFS on the complete tool graph. At each step, an LFM controller generates the next tool’s name and arguments, guided by the image, current tool-use trajectory, and matched examples. Once DFS is complete, the tool-use trajectory is determined, and then used to generate the query and answer.}
   \label{fig:method}
\end{figure*}

\noindent{\textbf{Tool-use VQA Task.}} Given the image scenario $\mathcal{I}$, the toolset $\mathcal{T}$, and the user query $\mathcal{Q}$,  a tool agent is required to provide an answer to the query $\mathcal{A}$ and the tool-use trajectory $\mathcal{P}=[(t_i, a_i)]_{i=1}^{n}$, in which $t_i\in \mathcal{T}$ represents the tool used in step $i$ and $a_i$ represents the calling arguments for $t_i$.
Consequently, each sample is a five-tuple $(\mathcal{I}, \mathcal{T}, \mathcal{Q}, \mathcal{A}, \mathcal{P})$. 

\noindent{\textbf{Toolset and Image Source.}}
We collect 10 tools with various functions as our toolset, including (1) Perception: ImageCaption, OCR, ObjectDetection, and RegionDescription. (2) Operation: DrawBox, GoogleSearch. (3) Logic: Calculator, Plot, ItemCount. (4) Creativity: TextToImage. These tools cover a wide range of real-world tasks and offer strong generalization capabilities (\eg GoogleSearch can retrieve any external knowledge).
The different sources of image $\mathcal{I}$ and the corresponding toolset $\mathcal{T}$ are shown in~\cref{fig:source}. These sources are diverse and complex, including real-world photography, e-commerce products, school exam charts, and other practical scenarios.

\subsection{ToolEngine}
\label{sec:construction}

We specify the definition of our real-world setting into the following criteria:
(1) Align with human users: Each sample's scenario \(\mathcal{I}\) and query \(\mathcal{Q}\) should align with real-world user needs. We have filtered out uninformative scenarios \(\mathcal{I}\), but it remains necessary to ensure that the tool-use queries \(\mathcal{Q}\) correspond to human demands.
(2) Real-world deployed tools: Real-world tool results are complex distributions with noise, and the model must learn how to deal with the noise from real results. Some work~\cite{patil2023gorilla, mialon2023gaiabenchmarkgeneralai} uses simulated tool outputs to reduce noise, but we believe this reduces the realism of the training environment.
(3) Multi-step reasoning: The tool-use trajectory $\mathcal{P}$ should involve multiple steps. Each step in it must have an inherent logical connection, forming a complete problem-solving trajectory. It should not include ineffective steps (\eg apply OCR to an image without text), or contain non-related tool usage in a single task (\eg use ``culinary world'' and ``NVIDIA’s stock'' sequentially, which should be split into two separate tasks).
We propose ToolEngine, an automated data synthesis pipeline that ensures these conditions. \cref{fig:method} illustrates the three core components of ToolEngine:

\noindent \textbf{Real-World Example Construction.}
To better reflect real user needs, we incorporate human prior tool-use knowledge into the LFM controller through in-context examples. %
Human-constructed examples provide prior knowledge about typical tool-use scenarios (\eg images containing specific entities may require a search tool, while tables with multi-digit numbers may necessitate a calculator). This information is crucial for synthesizing real-world queries.
Therefore, we ask human users to construct a small set of real-world examples $\mathcal{E} = \{(\mathcal{I}, \mathcal{T}, \mathcal{Q}, \mathcal{A}, \mathcal{P})\}$. Each example in this set captures a logically valid trajectory, serving as prior knowledge for automatic construction. 
This approach not only aligns with how humans intuitively associate tools but also provides a structured, intuitive foundation for the construction process. 

\noindent \textbf{Image-guided DFS on Tool Graph.} 
To construct a multi-step tool-use trajectory, we employ image-guided Depth-First Search (DFS) on a tool graph to construct a reasonable tool-use trajectory for each image. 
While LFM can directly generate reasoning sequences, unrealistic tool invocations often lead to less reliable tool-use trajectories in practice.
Our DFS trajectory involves real tool calls to extract detailed information from the input image, which is then used to generate challenging question-answer pairs.  
Specifically, we utilize the most advanced LFM (ChatGPT-4o-latest), denoted as $\mathcal{M}$, as the controller to select the tool at each step and generate the corresponding arguments. Formally: 
\begin{equation*}
    \resizebox{0.48\textwidth}{!}{$
    \begin{aligned}
        t_i &= \mathcal{M}(choices=\mathcal{T}, image=\mathcal{I}, examples=\text{Ret}(\mathcal{E}, \mathcal{P}_{i-1}))) \\
        a_i &= \mathcal{M}(tool=t_i, image=\mathcal{I}, examples=\text{Ret}(\mathcal{E}, \mathcal{P}_i)))
    \end{aligned}
    $}
\end{equation*}

\begin{figure}[t]
    \centering
    \includegraphics[width=1.0\linewidth]{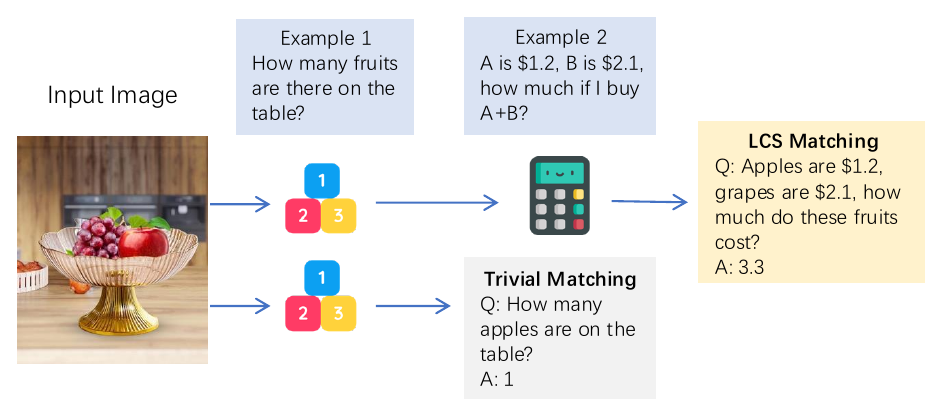}
    \caption{Comparison between different matching strategies. When a fixed example limits the diversity of generation, LCS matching can integrate multiple examples to enhance generation.}
    \label{fig:lcsteaser}
    \vspace{-0.5em}
\end{figure}

\noindent \textbf{LCS-based Example-Matching.} 
We combine the information obtained from different tools into a reasonable reasoning trajectory to perform multi-step reasoning.
Previous DFS-based methods typically match a fixed group of examples, which limited their ability to combine sufficient information for multi-step reasoning. As illustrated in ~\cref{fig:lcsteaser}, we propose an example-matching method based on the Longest Common Subsequence (LCS) algorithm, which enables dynamic matching during DFS. This approach better leverages the information extraction capabilities of diverse examples, facilitating more effective multi-step reasoning.
Specifically, in step $i$, the current solution process is denoted as $\mathcal{P}_i = [(t_1, a_1), (t_2, a_2), \cdots, (t_i, a_i)]$ and the set of example solution processes as $\mathcal{P}_{e}\in\mathcal{E}$. At each step, we perform the LCS matching between $\mathcal{P}_i$ and each element of $\mathcal{P}^{e}$, retrieving the top-k examples in $\mathcal{P}^{e}$ with the highest matching degree. Formally: 
$$
\text{Ret}(\mathcal{E}, \mathcal{P}_i) = \operatorname{TopK}_{\mathcal{P}^e \in \mathcal{E}} \Big\{ \operatorname{LCS}(\mathcal{P}^e, \mathcal{P}_i)  \Big\}
$$
This approach enables us to adaptively match each image with its most relevant example, while iteratively refining the match as new information is acquired at each step.

\noindent \textbf{Human Annotations and Analysis.} 
After the data synthesis process, we invite 10 human annotators to filter the generated data and re-annotate the highest-quality subset, resulting in 2,550 test samples. 
The initial synthetic data is estimated to have a 90.8\% accuracy in human validation on 4k randomly sampled data, which is higher than that reported in previous synthetic data methods~\cite{ma2024mms, gao2025multimodalagenttuningbuilding}.
Notably, the ChatGPT-4o-latest model, which we use to generate questions, demonstrates a low accuracy rate (less than 40\%) in solving them. 
This highlights the effectiveness of our ToolEngine pipeline in decoupling complex reasoning into a series of simpler, single-step tasks. While ChatGPT-4o-latest excels in handling $\mathcal{I}\rightarrow\mathcal{P}\rightarrow \mathcal{Q,A}$ process for straightforward, step-by-step reasoning, it faces challenges with the more challenging end-to-end $\mathcal{I,Q}\rightarrow\mathcal{P}\rightarrow\mathcal{A}$ reasoning. This observation indicates a clear gap between LFM's single-step and multi-step reasoning capabilities. %

\subsection{Dataset Statistics}
\label{sec:statistic}

\begin{table*}[ht]
\centering
\begin{minipage}{0.25\textwidth}
    \centering
    \small
    \begin{tabular}{ll}
    \toprule
    Statistics & Number \\
    \midrule
    Samples & 23655 \\
    Tool Calls & 65785 \\
    Text answers & 15806 \\
    Image answers & 7849 \\
    Query length & 15.74 \\
    Answer length & 2.69 \\
    Trajectory length & 2.78 \\
    \bottomrule
    \end{tabular}
    \centering
    \caption{Basic statistics.}
    \label{tab:sta}
\end{minipage}%
\begin{minipage}{0.31\textwidth}
    \centering
    \includegraphics[width=1.0\textwidth]{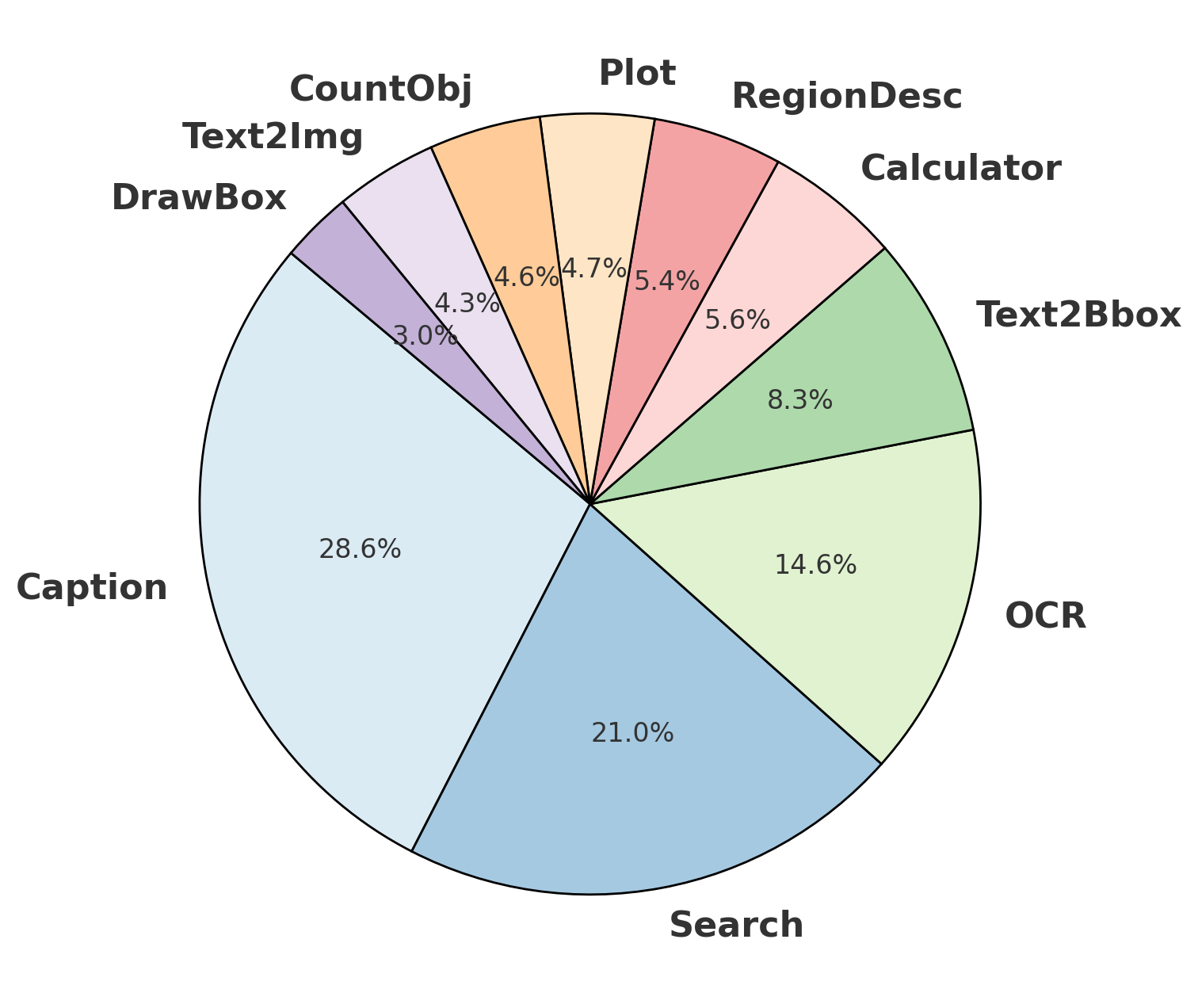}
    \vspace{-2em}
    \captionof{figure}{Tool frequency.} %
    \label{fig:frequency}
\end{minipage}
\begin{minipage}{0.36\textwidth}
    \centering
    \includegraphics[width=1.0\textwidth]{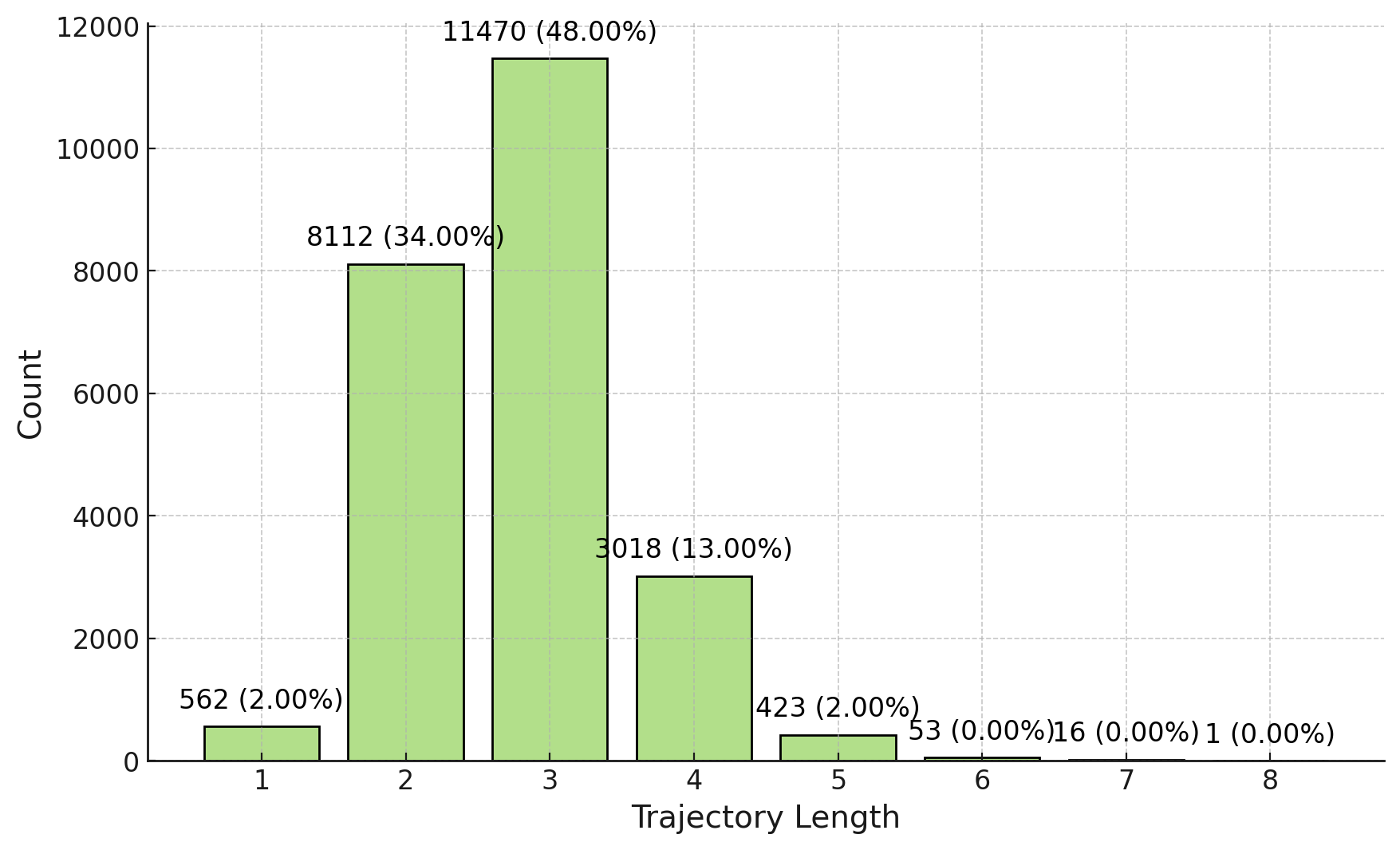}
    \captionof{figure}{Trajectory length distribution.} %
    \label{fig:distribution}
\end{minipage}%
\end{table*}

\cref{tab:sta} presents the basic statistics of ToolVQA, which encompasses various question types (with both text and image as answers) and complex reasoning processes (with an average reasoning trajectory length of 2.78 steps).  
Notably, ToolVQA's questions and answers are concisely formulated (with an average token count of 15.74 for questions and 2.69 for answers). In contrast, previous synthetic datasets~\cite{ma2024mms, gao2025multimodalagenttuningbuilding} often fail to control answer length, leading to hallucinated content within the answers, which is detrimental to training.  
\cref{fig:frequency} demonstrates that our tool frequency aligns with real-world user needs, reinforcing the real-world setting of our dataset. \cref{fig:distribution} presents the distribution of reasoning trajectory lengths in ToolVQA.

\begin{table}[h]
    \centering
    \small
    \begin{tabular}{lcccc}
    \toprule
    Method & Acc.$\uparrow$ & Cor. $\uparrow$ & Nec. $\uparrow$ & R.C. $\uparrow$\\
    \midrule
    ToolEngine & 90.8 & 85.2 & 87.51 & 2.38\\
    \midrule
    -w/o Example + LCS & 27.3 & 77.6 & 21.04 & 1.1 \\
    -w/o LCS & 41.6 & 81.4 & 54.26 & 1.61\\
    \bottomrule
    \end{tabular}
    \caption{Ablation study on different synthetic methods.}
    \label{tab:user}
\end{table}

\subsection{Ablation on ToolEngine}

To verify whether ToolEngine truly improves the quality of synthetic data, we conducted an ablation study on different synthetic methods. We invite 10 users to evaluate 1k randomly sampled data based on four key metrics:
(1) answer correctness (\textbf{Acc.}); 
(2) correlation (\textbf{Cor.}): whether each sample's image content and Q-A pair are aligned; 
(3) tool redundancy (\textbf{Nec.}): whether each tool call is necessary for obtaining the answer; 
and (4) reasoning complexity (\textbf{R.C.}): the average reasoning depth for each task (same as defined in~\cref{tab:comparison}).
The results, presented in \cref{tab:user}, show that Real-World Example and LCS matching significantly improved Acc., Nec., and R.C., which highlights their role in the DFS-based simulation of reasoning chains:  
(1) Real-World Example and LCS matching introduce prior knowledge aligned with image content, enhancing both answer accuracy and tool necessity.  
(2) Real-World Example and LCS matching incorporate different types of knowledge at each step, increasing the reasoning complexity.  
Furthermore, the absence of LCS matching has a significant impact on data quality (\eg Acc. dropped from 90.8 to 41.6), indicating that a fixed set of examples cannot adapt to different scenarios. Dynamically matching appropriate examples is crucial for significantly enhancing the generalization ability of the synthetic pipeline.
However, both methods had a limited impact on Cor., likely because the existing controller already effectively captures image content, minimizing errors caused by misleading examples.

\section{Experiments}
\label{sec:experiment}

\subsection{Training and Test Settings}
\label{sec:setting}

\noindent{\textbf{Training Objective.}} In our dataset $\mathcal{D}$, for each datapoint $\mathcal{E}=(\mathcal{I}, \mathcal{T}, \mathcal{Q}, \mathcal{A}, \mathcal{P})$ contains $n$ rounds of dialogue ($n-1$ rounds for tool usage and one round for final answer), we train our tool agent using cross-entropy loss:
$$
\mathcal{L} = \mathbb{E}_{\mathcal{E} \sim \mathcal{D}}\left[\frac{1}{n}\sum_{i=1}^{n} -\log p\Bigl(t_i, a_i, r_i \mid \mathcal{I}, \mathcal{T}, \mathcal{Q}, \mathcal{P}_{i-1}\Bigr)\right]
$$

\noindent{\textbf{Training Setting}.}
We divide ToolVQA into training and test sets: the training set consists of 21,105 automatically generated samples (with a human verification accuracy of 90.8\%), while the test set contains 2,550 human-reannotated samples.  
We fine-tune LLaVA-7B~\cite{liu2023visualinstructiontuning} in the Lego Agent framework~\cite{agentlego}. 

\noindent{\textbf{Test Setting}.}
We apply two test modes: end-to-end and step-by-step modes, to comprehensively evaluate LFMs' performance under different settings. The end-to-end mode only focuses on the end-to-end VQA ability of LFMs. Formally, we evaluate the $\mathcal{I}+\mathcal{Q}\rightarrow \mathcal{A}$ process. The step-by-step mode focuses on evaluating LM's tool-usage capabilities, formally the $\mathcal{I}+\mathcal{Q}+\mathcal{P}_{i}\rightarrow \mathcal{P}_{i+1}$ process. We use four metrics~\cite{wang2024gtabenchmarkgeneraltool}: (1) InstAcc (\textbf{Inst.}): the percentage of steps executed without errors; (2) ToolAcc (\textbf{Tool.}): the accuracy of tool selection; (3) ArgAcc (\textbf{Arg.}): the accuracy of tool's argument name prediction, and (4) SummAcc (\textbf{Summ.}): the accuracy of final answer's summary. 
Previous work has focused on the research of large language model (LLM)-based tool agents while ignoring vision-language model (VLM)-based tool agents and the end-to-end VQA abilities of VLMs. To bridge these gaps, our test involves three settings: (1) \textbf{VLM}, directly answering questions end-to-end, formally $\mathcal{Q}+\mathcal{I}\rightarrow \mathcal{A}$, (2) \textbf{VLM+tool}, calling the tool through VLM, formally $\mathcal{Q}+\mathcal{I}+\mathcal{T}\rightarrow \mathcal{P}+\mathcal{A}$, (3) \textbf{LLM+tool}, calling the tool through LLM, formally $\mathcal{Q}+\mathcal{T}\rightarrow \mathcal{P}+\mathcal{A}$ (Note that image information can still be obtained through tools such as caption under this setting).

\subsection{Main Results } 
\label{sec:mainres}

\begin{table*}[ht]
\centering
\small
\begin{tabular}{lcccccccc}
\toprule
\multirow{2}{*}{Model} & \multirow{2}{*}{Setting} & 
        \multicolumn{3}{c}{\textbf{End-to-End Mode}} & 
        \multicolumn{4}{c}{\textbf{Step-by-Step Mode}} \\
& & {\makecell[c]{Acc. \\ (\%)}}  & {\makecell[c]{Tool Call \\ (Times)}} & {\makecell[c]{Tool Error \\ (Times)}} & {\makecell[c]{Inst. \\(\%)}} & {\makecell[c]{Tool.\\(\%)}} & {\makecell[c]{Arg.\\(\%)}} & {\makecell[c]{Summ.\\(\%)}} \\
\midrule
\rowcolor{cyan!20} \multicolumn{9}{l}{\textbf{\textit{Closed-source LFMs}}} \\
ChatGPT-4o-latest  
& VLM & \textbf{38.29} & - & - & - & - & - & - \\
& VLM + tool & 34.96 & 2003 & 1804 & 36.5 & 14.68 & 8.92 & 56.1 \\
 & LLM + tool & 37.62 & 4687 & 1962 & 73.02 & 41.12  & 25.27 & 65.45 \\
Claude-3-5-sonnet  
& VLM & 30.33 & - & - & - & - & - & - \\
& VLM + tool & 30.52 & 585 & 524 & 45.71 & 16.94 & 7.36 & 46.06 \\
 & LLM + tool & 24.48 & 1834 & 672 & 66.83 & 39.29 & 13.98 & 75.52 \\
GPT-3.5-Turbo & LLM + tool & 18.37 & 3178 & 1321 & 73.24 & 30.46 & 20.08 & 58.18 \\
\midrule
\rowcolor{orange!20} \multicolumn{9}{l}{\textbf{\textit{Open-source LFMs}}} \\
Qwen2-VL-7B-Instruct 
& VLM & 3.64 & - & - & - & - & - & - \\ 
& VLM + tool & 5.86 & 1373 & 1051 & 35.12 & 9.02 & 1.09 & 38.18 \\
 & LLM + tool & 6.78 & 1366 & 1089 & 39.46 & 11.2 & 3.55 & 44.85 \\
Qwen2-7B-Instruct & LLM + tool & \textbf{11.53} & 1773 & 703 & 63.1 & 24.45 & 8.2 & 69.7 \\
LLaVA-v1.5-7B & VLM & 8.57 & - & - & - & - & - & - \\
& VLM + tool & 1.17 & 9684 & 9684 & 16.39 & 9.43 & 0 & 0.01 \\
& LLM + tool & 1.66 & 10421 & 10421 & 9.25 & 15.03 & 0 & 11.52 \\
LLaMA-3-8b-instruct & LLM + tool & 4.93 & 6607 & 6542 & 51.63 & 14.13 & 0.53 & 41.99 \\
\midrule
Qwen2-VL-2B-Instruct  
& VLM & \textbf{7.71} & - & - & - & - & - & - \\
& VLM + tool & 2.1 & 4067 & 3915 & 47.16 & 12.7 & 0.41 & 21.82 \\
 & LLM + tool & 2.22 & 6892 & 6655 & 36.34 & 10.52 & 0.68 & 32.12 \\
Qwen2-1.5B-Instruct & LLM + tool & 4.5 & 7933 & 6141 & 50.95 & 21.45 & 4.51 & 44.24 \\
\midrule
Tuned LLaVA-7B \textbf{(Ours)}
& VLM & 7.21 & - & - & - & - & - &  - \\
& VLM + tool & \textbf{18.8} & 4311 & 617 & 86.62 & 61.61 & 39.34 & 30.91 \\
& LLM + tool & 18.43 & 4462 & 680 & 87.51 & 62.3 & 40.57 & 26.67 \\
\bottomrule
\end{tabular}
\caption{Results on ToolVQA's test set.}
\label{tab:ours}
\end{table*}

\noindent \textbf{Scale.} As shown in \cref{tab:ours}, the tool-use capabilities of LFMs generally improve with the increase in model parameters. For example, Qwen2-7B-instruct surpasses Qwen2-2B-instruct by 7.03\%, and closed-sourced LM (larger than 100B) is generally better than open-sourced (2B,7B). After fine-tuning, our 7B model's performance (last 3 lines) is close to the large scale LFM GPT-3.5-Turbo. This demonstrates that fine-tuning significantly enhances the model's real-world tool-use capabilities, even in smaller models.

\noindent \textbf{Settings.} By comparing the performance of the same model under different settings (VLM, VLM+tool, and LLM+tool) in~\cref{tab:ours}, we identify several interesting observations.
(1) In models such as GPT-4o, Qwen2-7B/2B, and LLaVA-7B, the VLM+tool setting performs worse than the LLM+tool setting, suggesting that the visual modules in current LFM architectures are still ineffective at guiding tool usage.
(2) For models including GPT-4o, Claude-3.5, LLaVA-7B, and Qwen2-2B, we find that VLM+tool underperforms VLM alone. This indicates that the noise introduced by tool usage can outweigh the benefits brought by the tools.
However, the tuned LLaVA model demonstrates a clear progression: $\text{VLM} < \text{LLM+tool} < \text{VLM+tool}$, suggesting that fine-tuning can indeed enhance the model’s ability to make effective use of tools while suppressing tool-induced noise. Nevertheless, the similar performance of the LLM+tool and VLM+tool settings implies that the visual module's potential in facilitating tool use has not yet been fully explored.

\noindent \textbf{Bottlenecks.} The last two rows of~\cref{tab:ours} indicate that the performance of fine-tuned models is often constrained by their weakest capabilities, which are typically argument prediction and answer summarization. While fine-tuning significantly improves instruction formatting and tool selection, it makes little progress in weaker capabilities.  
Specifically, instruction formatting and tool selection can be accomplished through memorizing patterns, as the number of reasonable tool invocation types in real-world scenarios is limited. However, argument prediction and answer summarization require the model to genuinely understand the new information returned by tools and extract meaningful responses from it.  
This information varies across different scenarios and queries, and even within multi-turn dialogues of the same scenario. Consequently, a high level of generalization capability in vision-language joint reasoning is essential to handle these issues in tool-use tasks.

\subsection{Out-of-Distribution Benchmarks}
\label{sec:oodres}

\begin{table*}[h]
\centering
    \begin{tabular}{lcccccccc}
    \toprule
    Model  & {\makecell[c]{TextVQA~\cite{singh2019vqamodelsread}}} & {\makecell[c]{TallyQA~\cite{acharya2018tallyqaansweringcomplexcounting}}} & {\makecell[c]{InfoSeek~\cite{chen2023pretrainedvisionlanguagemodels}}} & {\makecell[c]{GTA~\cite{wang2024gtabenchmarkgeneraltool}}} & {\makecell[c]{TEMPLAMA~\cite{Dhingra_2022}}} \\
    \midrule
    GPT-3.5-Turbo  & 36.3 & \underline{61} & \underline{11.3} & \underline{23.62} & \textbf{33.67} \\
    LLaVA-7B  & \underline{41.2} & 60.1 & 5.2 & 12.12 & 3.06 \\
    Tuned LLaVA-7B  & \textbf{47} & \textbf{64.3} & \textbf{13.8} & \textbf{33.29} & \underline{21.43} \\
    \bottomrule
    \end{tabular}
    \caption{Results on OOD benchmarks.}
    \label{tab:ood}
\end{table*}
To test the generalization capability of our tuned model, we conduct experiments on out-of-distribution (OOD) benchmarks including (1) Seen tools, unseen queries: TextVQA~\cite{singh2019vqamodelsread}, TallyQA~\cite{acharya2018tallyqaansweringcomplexcounting}, InfoSeek~\cite{chen2023pretrainedvisionlanguagemodels}; and (2) Unseen tools: GTA~\cite{wang2024gtabenchmarkgeneraltool}, TEMPLAMA~\cite{Dhingra_2022}. %
The results are shown in the~\cref{tab:ood}. Our fine-tuned model significantly outperforms the baseline model LLaVA-7B by 5.8\%, 4.2\%, 8.6\%, 21.17\%, and 18.37\% accuracy on these datasets, showing excellent generalizability. Its performance also surpasses the large closed-source model GPT-3.5-Turbo on unseen queries. However, its performance on TEMPLAMA is inferior to GPT-3.5-Turbo, possibly because GPT encountered similar data during its training (due to its great coverage of extensive training data), whereas our model had no exposure to such data at all.

\begin{table*}[h]
\resizebox{0.46\textwidth}{!}{
\begin{minipage}{0.48\textwidth}
    \centering
    \begin{tabular}{lcccc}
    \toprule
    Model & Zero-shot & 1-shot & 5-shot & 10-shot \\
    \midrule
    GPT-4o & 34.96 & 37.20 & 38.41 & 38.63  \\
    GPT-3.5-Turbo & 18.37 & 19.83 & 20.40 & 19.04 \\
    LLaVA-7B & 1.17 & 3.45 & 4.27 & 3.67 \\
    Tuned LLaVA-7B & 18.80 & 19.41 & 21.13 & 20.69 \\
    \bottomrule
    \end{tabular}
    \caption{Few-shot performance on ToolVQA.}
    \label{tab:fewshot}
\end{minipage}
}
\resizebox{0.5\textwidth}{!}{
\begin{minipage}{0.56\textwidth}
    \centering
    \vspace{-0.1em}
    \begin{tabular}{lcccccccc}
    \toprule
    {Model} & {Setting} &  Acc. & {\makecell[c]{Tool Call}} & {\makecell[c]{Tool Error}} \\
    \midrule
    Tuned LLaVA-7B & VLM + tool & 18.8 & 4311 & 617 \\
    \midrule
    -w/o Caption & VLM + tool & 11.1 & 1861 & 192 \\
    -w/o $\mathcal{P}$ & VLM & 8.01 & - & - \\
    -w/o $\mathcal{P+I}$ & LLM & 0.18 & - & -  \\
    \bottomrule
    \end{tabular}
\caption{Ablation study on training methods.}
\label{tab:ablations}
\end{minipage}
}
\begin{minipage}{1.0\textwidth}
    \vspace{1.0em}
    \centering
    \includegraphics[width=0.9\linewidth]{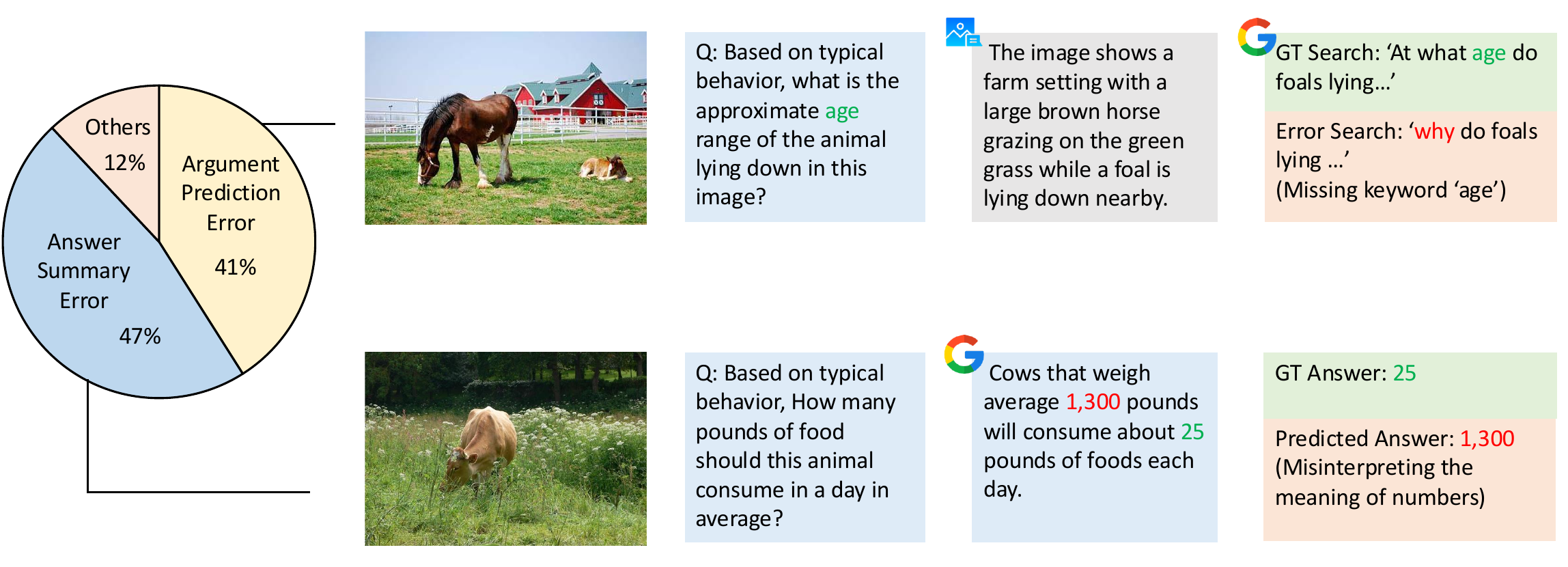}
    \captionof{figure}{Visualization of the main error types made by tuned LLaVA.}
    \label{fig:visualization}
\end{minipage}
\end{table*}

\subsection{Few-shot In-context Learning}
\label{sec:icl}

Few-shot in-context learning (ICL) has become a common and important technique in tool-use tasks. However, prior model fine-tuning work often overlooks evaluating the ICL capabilities of fine-tuned models. We believe that fine-tuning and ICL can be complementary, offering significant potential for improvement.
\cref{tab:fewshot} shows few-shot ICL performance on ToolVQA. GPT-4o steadily improves with more shots, while GPT-3.5 and LLaVA both show limited gains and even decline slightly at 10-shot. This may be due to long contexts impairing the LLM’s ability to interpret precise tool-use instructions. Fine-tuned LLaVA outperforms the baseline and continues to benefit from ICL. These results suggest that (1) compared to fine-tuning, ICL's improvement is limited and affected by context lengths, and (2) fine-tuned models can still benefit from ICL, highlighting the complementary value of these two approaches in enhancing tool-use performance.

\subsection{Ablation Study}
\label{sec:ablation}

To further evaluate the quality of the fine-tuning data from multiple perspectives, we conduct an ablation study of the different training strategies. As shown in~\cref{tab:ablations}, removing the ImageCaption tool significantly degrades both accuracy and parameter correctness. This highlights the model's reliance on high-quality captions for initial scene understanding, likely due to its limited inherent visual capability for fine-grained information extraction.
When all tools are removed and ToolVQA is reduced to a single-hop VQA model, performance drops substantially, underscoring the necessity of multi-step tool usage.
In the absence of both tools and images, performance approaches zero, indicating that the model is heavily dependent on visual input when answering questions in ToolVQA.

\subsection{Error Analysis}
\label{sec:visualization}

In \cref{fig:visualization}, we randomly select 100 errors made by tuned LLaVA, and manually count and classify the main error types. The two main error types are:  
(1) Argument prediction error, for example, missing essential keywords ``age'' during a search.  
(2) Answer summary error, for example, extracting incorrect information ``1,300'' from tool outputs.  
The common cause of these errors is the model’s failure to recognize key terms (\eg ``age'' and ``pound'') in the tool outputs, even though this should not be a challenging task for a language model.  
This suggests that the model still struggles with dynamically adapting to additional information provided by tools. Since multi-step reasoning often suffers from error accumulation (early errors propagate to subsequent steps), errors in processing tool results at any step will affect the results. Fine-tuning has limited improvement on the capability for dynamically processing tool results, but ToolVQA still provides a valuable benchmark for evaluating this capability. 

\section{Conclusion}
\label{sec:conclustion}
We present ToolVQA, a large multimodal dataset for real-world tool-augmented reasoning. Using our ToolEngine pipeline, we generate 23K real-world samples with implicit multi-step reasoning over 10 tools and 7 domains. This dataset challenges existing models beyond synthetic settings and better reflects real user needs. Experiments show that LLaVA-7B fine-tuned on ToolVQA not only excels on the in-domain test set but also outperforms GPT-3.5-turbo on five OOD benchmarks, establishing a new state-of-the-art for open-source models in complex tool-use VQA. ToolVQA serves as both a benchmark and a training ground for developing more capable, generalizable tool-using agents.

\noindent\textbf{Acknowledgements.}This work was supported by the grants from the National Natural Science Foundation of
China 62372014, Beijing Nova Program and Beijing Natural Science Foundation 4252040.

{
    \small
    \bibliographystyle{ieeenat_fullname}
    \bibliography{main}
}

\clearpage
\setcounter{page}{1}
\maketitlesupplementary

In this supplementary material, we provide more details of the paper. 
In~\cref{sec:supptool}, we introduce the specific functionalities and design principles of the ToolVQA toolset.
In~\cref{sec:suppalgo}, we provide the pseudocode of the LCS-based example matching algorithm that is applied in ToolEngine. 
In~\cref{sec:suppuser}, we provide more user study details, including the composition of our users, the sampling process of initial user logs, and methods for data quality evaluation metrics.
In~\cref{sec:suppexpr}, we provide more experiment details, including the training pipeline, training hyperparameters, evaluation setting, and evaluation metrics. 
In~\cref{sec:suppprompt}, we provide the detailed prompts used in our pipeline.
In~\cref{sec:suppdemo}, we provide some demonstrations of our real-world examples used to prompt ToolEngine, and some examples of ToolVQA's test set.

\section{Toolset}
\label{sec:supptool}

\begin{table*}[t]
  \centering
  \begin{tabular}{llccc}
    \toprule
    Tool Name & Description & Model & Input & Output \\
    \midrule 

\begin{minipage}[b]{0.04\columnwidth}
		\centering
		\raisebox{-.15\height}{\includegraphics[width=\linewidth]{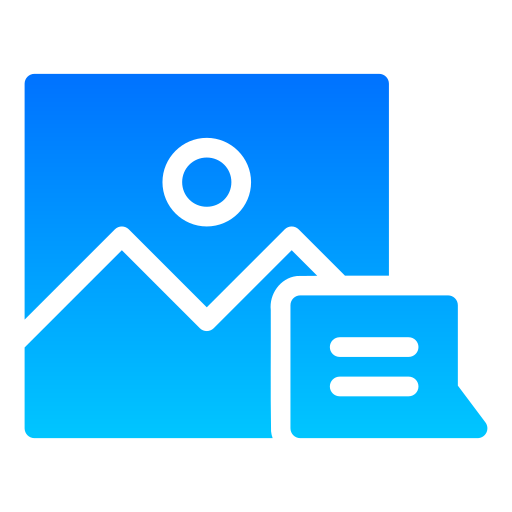}}
	\end{minipage} ImageCaption & Describe an image  & ChatGPT-4o-latest &  image & text \\

 \midrule 
 
  \begin{minipage}[b]{0.04\columnwidth}
		\centering
		\raisebox{-.15\height}{\includegraphics[width=\linewidth]{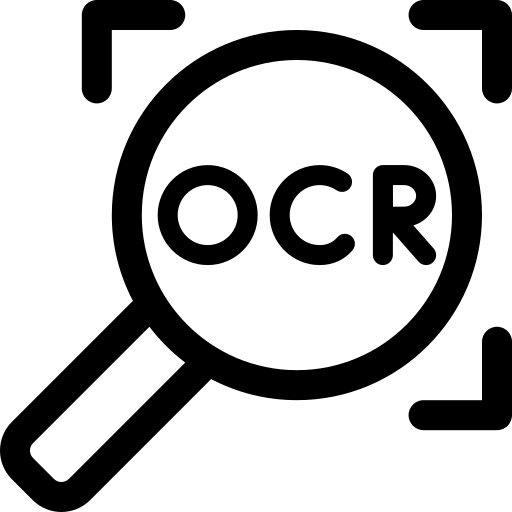}}
	\end{minipage} OCR & Recognize the text in image  & PaddleOCR~\cite{du2020ppocr} & image & text \\

\midrule 
 
  \begin{minipage}[b]{0.04\columnwidth}
		\centering
		\raisebox{-.15\height}{\includegraphics[width=\linewidth]{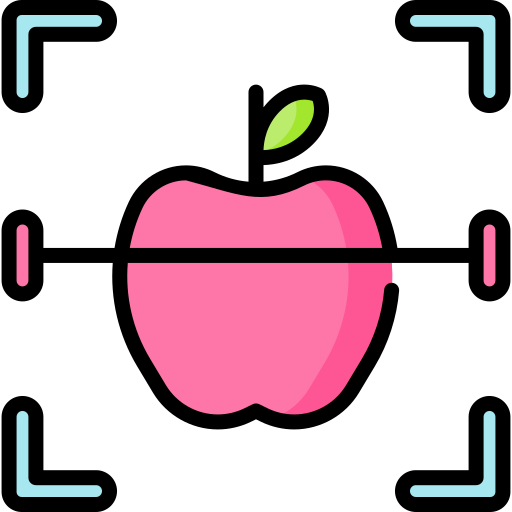}}
	\end{minipage} ObjectDetection & Detect an object in image  &  MM-Grounding-DINO~\cite{zhao2024opencomprehensivepipelineunified} & image, object & bbox \\
\midrule

 \begin{minipage}[b]{0.04\columnwidth}
		\centering
		\raisebox{-.15\height}{\includegraphics[width=\linewidth]{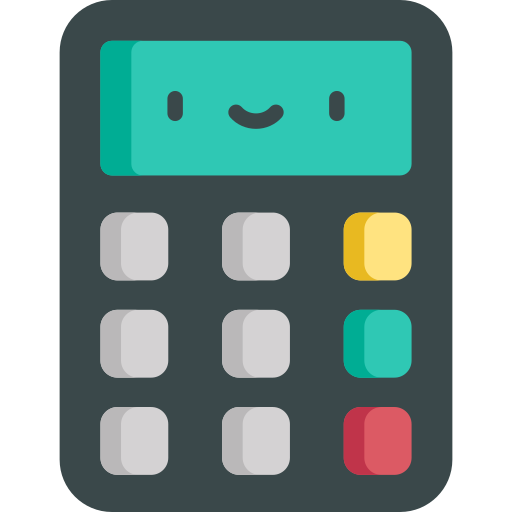}}
	\end{minipage} Calculator & Calculate a math expression & Python.math & expression & number \\

 \midrule

 \begin{minipage}[b]{0.04\columnwidth}
		\centering
		\raisebox{-.15\height}{\includegraphics[width=\linewidth]{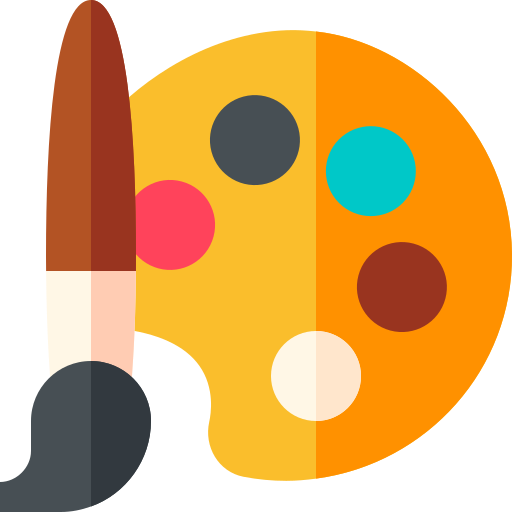}}
	\end{minipage} TextToImage & Generate an image based on text & Stable-Diffusion-v1.5~\cite{rombach2021highresolution} & text & image \\

\midrule

     \begin{minipage}[b]{0.04\columnwidth}
		\centering
		\raisebox{-.15\height}{\includegraphics[width=\linewidth]{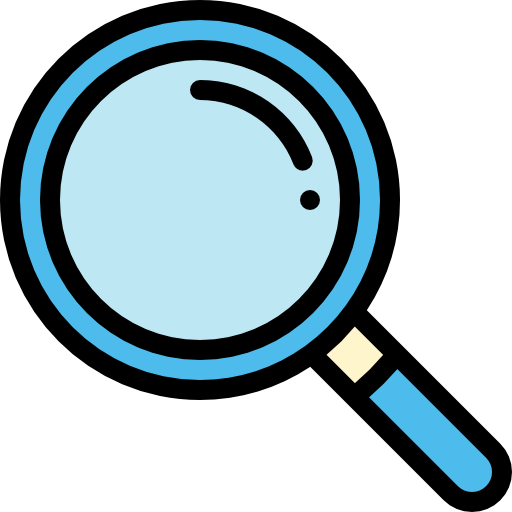}}
	\end{minipage} RegionDescription & Describe attribute for a region  & ChatGPT-4o-latest & image, bbox, attribute  &  text\\

\midrule

     \begin{minipage}[b]{0.04\columnwidth}
		\centering
		\raisebox{-.15\height}{\includegraphics[width=\linewidth]{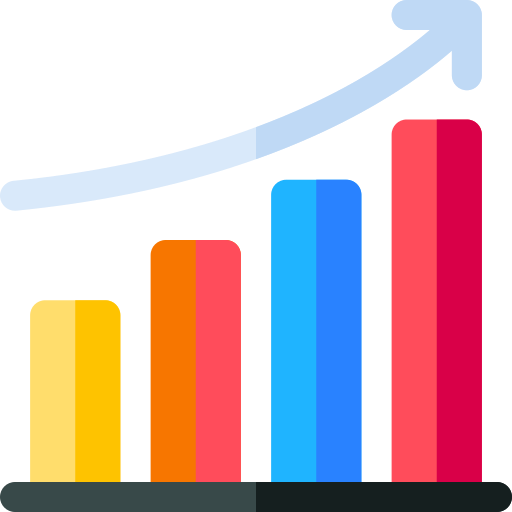}}
	\end{minipage} Plot & Plot a diagram  & Python.matplotlib~\cite{Hunter:2007} & python code & image \\

\midrule

     \begin{minipage}[b]{0.04\columnwidth}
		\centering
		\raisebox{-.15\height}{\includegraphics[width=\linewidth]{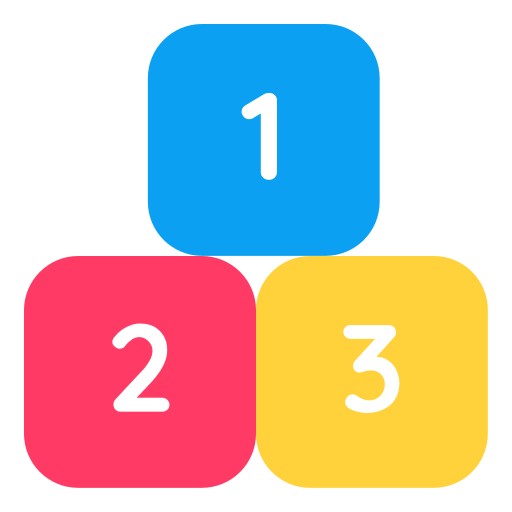}}
	\end{minipage} ItemCount & Count the number of an object & ChatGPT-4o-latest & image, object & number \\

\midrule

     \begin{minipage}[b]{0.04\columnwidth}
		\centering
		\raisebox{-.15\height}{\includegraphics[width=\linewidth]{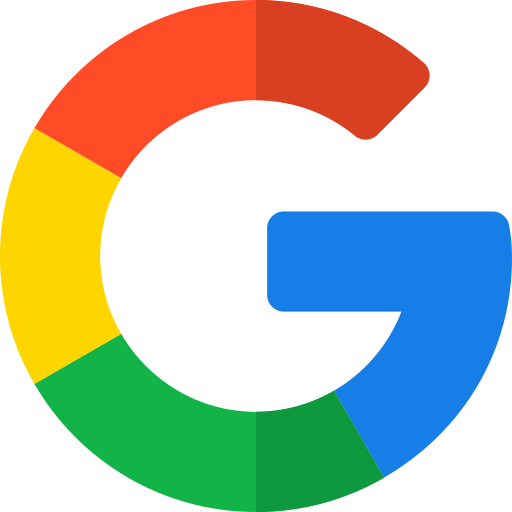}}
	\end{minipage} GoogleSearch & Search external knowledge  & Google Serper API & query & text \\

\midrule

     \begin{minipage}[b]{0.04\columnwidth}
		\centering
		\raisebox{-.15\height}{\includegraphics[width=\linewidth]{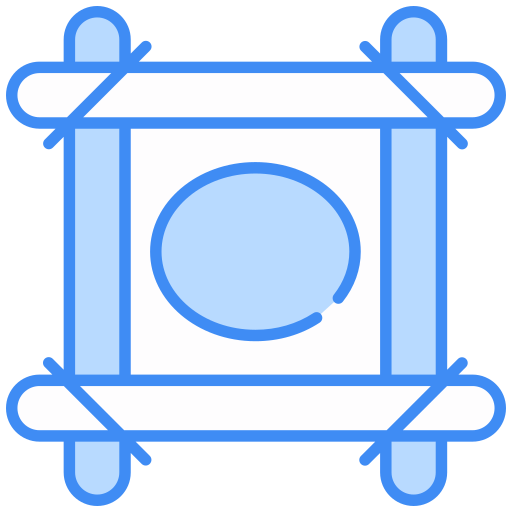}}
	\end{minipage} DrawBox & Draw a box on image  & Python.pillow & image, bbox & image \\

    \bottomrule
  \end{tabular}
  \caption{Details of our toolset.}
  \label{tab:supptool}
\end{table*}

\cref{tab:supptool} shows the details of our toolset. We adopt an open-sourced library AgentLego~\cite{agentlego} to build our toolset and adopt Lagent~\cite{lagent} framework to let LFM-based tool agents interact with the tools.

Although our toolset is not very large, these tools are all selected to address LFM's shortcomings in certain aspects, such as external knowledge acquisition, text recognition, image generation, etc. Unlike previous works~\cite{qin2023toolllm, ma2024mms} that set up a tool for each sub-task, our tools have diverse capabilities and strong generalization. For example, Google Search can search for external knowledge such as news, historical events, data, public information, academic papers, etc. As we have discussed in Sec.1 of the main paper, binding single-function tools to specific types of problems will turn the tool usage task into a query pattern recognition task, which does not allow LFMs to gain an intrinsic understanding of the tool affordance and functionality. In contrast, Our highly generalized tools contain countless possible argument combinations. Using them well requires a comprehensive understanding of the tools, scenarios, and queries, which is more similar to human tool use.

In order to ensure that the controller can see the image during the whole process of asking questions, we need to send the image to each conversation of the controller. However, this requires a high cost. To improve efficiency, we change it to fix the first tool to ImageCaption/OCR when asking questions, so that the controller can understand the overall information of the image in the first step. At the same time, this can also provide the necessary basic information for LFMs when answering queries. The ablation experiment in Sec.4.3 of the main paper shows that Caption is important for the fine-tuned model to answer queries.

\section{Algorithm}
\label{sec:suppalgo}

\begin{algorithm}[H]
\caption{Select Top-k Examples by LCS}
\begin{algorithmic}[1]
\Function{LCS}{$A$, $B$}
    
    \For{$i$ from $1$ to $|A|$}
        \For{$j$ from $1$ to $|B|$}
            \If{$A[i-1] == B[j-1]$}
                \State \texttt{dp}$[i][j] = $ \texttt{dp}$[i-1][j-1] + 1$
            \Else
                \State \texttt{dp}$[i][j] = \max($\texttt{dp}$[i-1][j], $ \texttt{dp}$[i][j-1])$
            \EndIf
        \EndFor
    \EndFor
    \State \Return \texttt{dp}$[|A|][|B|]$
\EndFunction
\State
\State Initialize an empty list \texttt{LCS\_values}
\State Initialize a 2D array \texttt{dp} of size $(|A|+1) \times (|B|+1)$ with all values 0
\For{each $\mathcal{P}_{ej}$ in $\mathcal{P}_e$}
    \State Calculate \texttt{LCS\_length} = LCS($\mathcal{P}_i$, $\mathcal{P}_{ej}$)
    \State Append $(\mathcal{P}_{ej}, \texttt{LCS\_length})$ to \texttt{LCS\_values}
\EndFor
\State \Return Ret = Top-k(\texttt{LCS\_values})

\end{algorithmic}
\label{alg:lcs}
\end{algorithm}

\cref{alg:lcs} shows the pseudocode for our LCS-based example matching algorithm introduced in Sec.3.2 of the main paper. The LCS algorithm is a classic method for measuring the similarity between two ordered lists. Unlike similarity calculation modules trained using neural networks, LCS offers a simple yet effective solution that ensures both accuracy and efficiency. Moreover, LCS is well-suited for handling longer trajectories in future application scenarios, maintaining its robustness and precision even as complexity increases. 

\section{User Study Details}
\label{sec:suppuser}

We randomly select 10 real-world users from universities, including both students and professors from different disciplines. The participants spanned five fields: Mathematics, Computer Science, Economics, Chinese Language, and Art, with two users from each discipline.

\subsection{Toolset Selection}

\begin{figure}[h]
    \centering
    \includegraphics[width=1.0\linewidth]{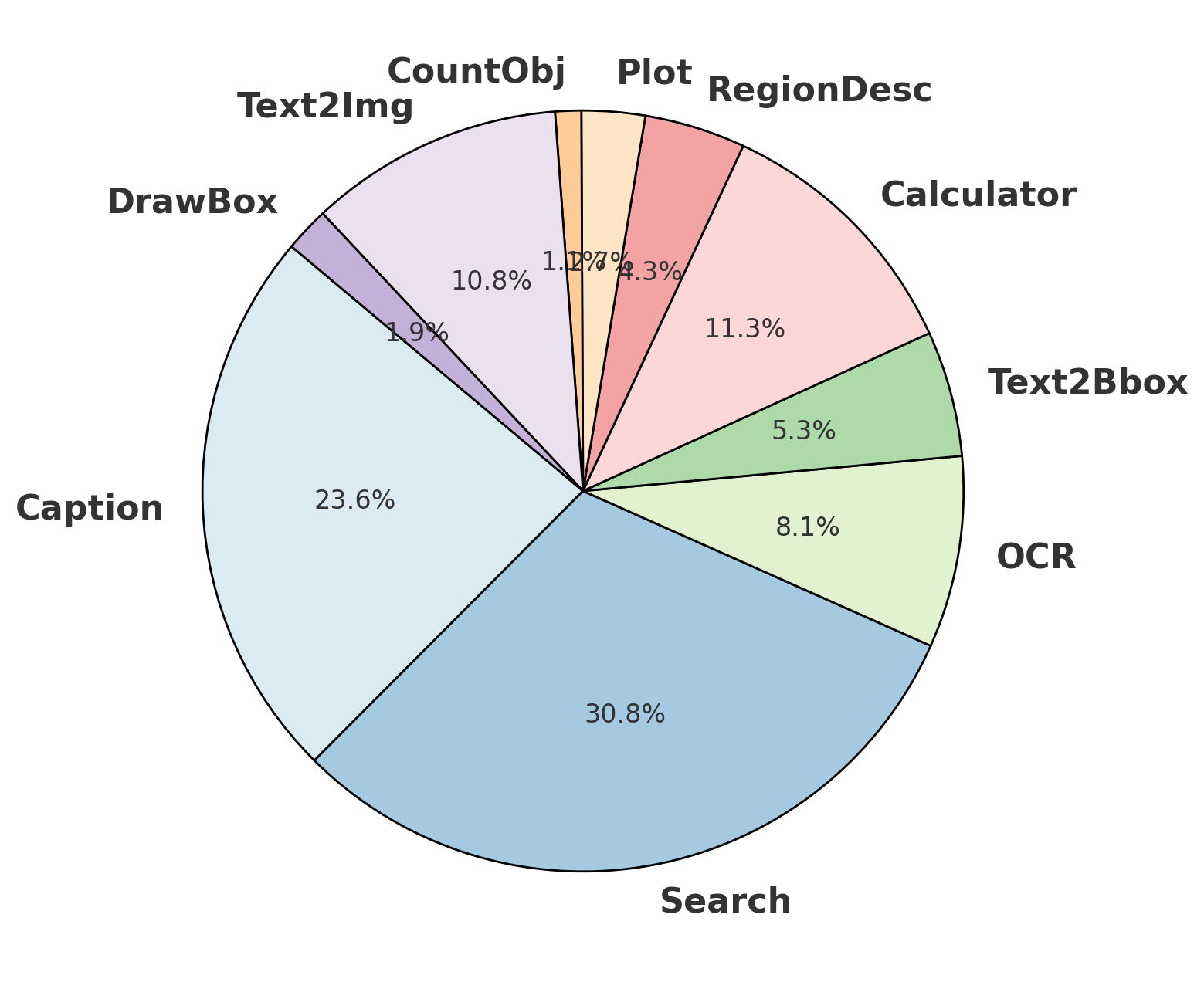}
    \caption{Real-world users tool frequency.}
    \label{fig:tool-freq-supp}
\end{figure}

We invite users to document 15 common tool-use scenarios they frequently encounter in their work, along with the corresponding trajectories. We then merged functionally similar tools and selected the 10 most frequently occurring tools as our final toolset. The detailed frequency distribution is shown in~\cref{fig:tool-freq-supp}. Subsequently, human experts discussed and consolidated similar scenarios and trajectories. Through this process, we refined the initial 150 scenarios into a final set of 34 representative examples. These examples cover all 10 tools and most reasonable tool combinations, with only some differences in the number of tool usages (e.g., counting the number of different objects, iterative searching for external knowledge) that need to be compensated by our LCS-based example matching algorithm. These queries go through multiple rounds of iteration and expert discussion, aiming to meet the following requirements: (1) trajectories are necessary to answer queries; (2) understanding image is necessary to answer queries; (3) all queries cover the vast majority of reasonable trajectories; and (4) queries are helpful to real human life. 

\subsection{Evaluation Metrics}

For each dataset, we randomly sampled 100 samples and invite users to annotate them.
To calculate the metrics of Acc., Corr., and Nec.. on our dataset, we ask users to individually review each image-question-answer pair along with its corresponding trajectory. For deterministic-answer metrics (Acc., Nec..), we design a binary (Yes/No) selection. For degree-based metrics (Corr.), we use a 1-10 scoring scale to capture nuanced differences.
To calculate the Reasoning Complexity (R.C.), we ask users to write a tree-of-thought (ToT)~\cite{yao2023treethoughtsdeliberateproblem} to solve the query for each sample. Then we calculate the mean depth of the samples' ToTs.

\section{Experiment Details}
\label{sec:suppexpr}

\subsection{Training}
\label{sec:supptrain}

We use XTuner~\cite{2023xtuner} as our training framework. We finetune LLaVA-7B using LoRA~\cite{hu2021lora} algorithm, with $\text{batch\_size}=2, \text{learning\_rate}=2e-4$ on 4xGTX3090 GPUs for 4000 epochs.

\subsection{Evaluation}
\label{sec:suppeval}

We use OpenCompass~\cite{2023opencompass} as our evaluation framework. On our ToolVQA test set, we follow all the metrics used in GTA~\cite{wang2024gtabenchmarkgeneraltool} that are mentioned in Sec.4.1 of the main paper. When comparing the output with the ground truth, we follow GTA to divide the ground truth into a whitelist and a blacklist for matching. \cref{fig:testset} demonstrates some examples of our test set. The whitelist and blacklist are manually labeled and can generally accept most of the correct answers. We only evaluate 1622 samples with text answers under end-to-end mode due to the lack of generation metrics. We evaluate all 2550 samples under step-by-step mode.

On the various out-of-distribution (OOD) benchmarks (TextVQA~\cite{singh2019vqamodelsread}, TallyQA~\cite{acharya2018tallyqaansweringcomplexcounting}, InfoSeek~\cite{chen2023pretrainedvisionlanguagemodels} and GTA~\cite{wang2024gtabenchmarkgeneraltool}), we use their own evaluation metrics. Since the time required for tool-use VQA is significantly longer than that of traditional VQA (the former is about 4-6 times that of the latter, limited by the running speed of the tool itself), we follow previous works~\cite{castrejon2024hammrhierarchicalmultimodalreact, suris2023vipergpt} to randomly sample 1,000 examples from their test set for testing to ensure fairness and accuracy as much as possible. In addition, we notice that LLaVA~\cite{liu2024improvedbaselinesvisualinstruction} provide additional OCR tokens in each image to help VLM answer when testing TextVQA, and we believe that this cannot accurately evaluate VLM's text recognition ability, so we follow previous work~\cite{liu2024ocrbenchhiddenmysteryocr} and remove these OCR tokens in the evaluation, which made our test results significantly lower than the results in~\cite{liu2024improvedbaselinesvisualinstruction}.

\section{Prompts}
\label{sec:suppprompt}

We provide the exact prompts that we used in our pipeline.

\subsection{ToolEngine}

In the ToolEngine, we use the LFM-based controller (in the main paper Fig.3) to construct the multi-step tool-use VQA samples. The controller has three main purposes: (1) select which tool to use in the next step; (2) explain why we select that tool (to obtain the chain-of-thought of the sample); (3) come up with the final question and answer of the sample. We list the prompt examples of each part below:

(1) Select the next tool:

\begin{lstlisting}
You are a smart tool selector. I will provide you with some information extracted from an image, along with a list of available tool options for the next steps. Please choose the most suitable tool to obtain more information or generate a new image. 

The tool options are as follows: 
{options} 

Your response should consist of two parts:

1. **Thought:** Explain your reasoning behind how you decide to choose the next tool.
2. **Choice:** Provide the specific tool you have selected.

Here are some examples to help you understand the task.

{examples}

Now that you understand the approach and format for selecting tools, I will provide you with the necessary information. Please choose the next tool using the same format.

Information:
{context}
\end{lstlisting}

(2) Explain the reason:

\begin{lstlisting}
You are a smart information processor. I will provide you with a problem, an answer, and a process for solving the problem using different tools. Your task is to describe the thinking behind solving the problem, specifically explaining the purpose of using each tool.

Your response should include several lines, one for each tool, and each line should contain two parts:

1. **Tool Name:** The name of the tool used.
2. **Thought:** Explain the purpose of using the tool, including the information you expect to get from it to solve the problem.

Here are some examples to guide you:

Example 1:
```
{example1}
```

Example 2:
```
{example2}
```

Now that you understand the format, I will provide you with the information. Please generate your response accordingly.

Question: {question}

Solving Process:
{context}

Answer: {answer}
\end{lstlisting}

(3) Come up with the final question and answer:

\begin{lstlisting}
You are a question generator that creates valuable queries based on extracted information from a tool process. The task is to formulate questions about the image data that meet the following conditions:

1. The answer must be the result returned by the LAST tool call. If the answer is a long sentence, you need to summarize it into a single word or phrase.

2. The question should address a scenario that can occur in real-life situations and meet practical needs.

3. It must be solvable through the tool call, avoiding trivial or overly complex inquiries unrelated to the data.

4. The answer requires analyzing the image or information extracted from the image. The direct content of the image must not appear in the question. Instead, refer to the image as "this," "image," or "picture."

Your response should consist of three parts:

1. **Thought:** Your reasoning for generating the question.
2. **Question:** The question you asked.
3. **Answer:** The answer to the question, either text or a picture generated by the last tool.

Here are some examples to help clarify:

{examples}

Now, I will provide you with the process information. Please create your answer accordingly.

Process: 
{context}
\end{lstlisting}

\subsection{Fine-tuning}

An example of the fine-tuning prompts is shown below: 
(The question of this example is ``In the image, Where can someone buy the soda shown in this image?'')

\begin{lstlisting}
[system] You are an assistant who can utilize external tools.
[
    {
        "name": "ImageDescription", 
        "description": "A useful tool that returns a brief description of the input image.", 
        "inputs": 
        [{
            "type": "image", 
            "name": "image", 
            "description": null, 
            "optional": false, 
            "default": null, "filetype": null
        }], 
        "outputs": 
        [{
            "type": "text", 
            "name": null, 
            "description": null, 
            "optional": false, 
            "default": null, "filetype": null
        }]
    }, 
    {
        ... (other tools)
    }, 
    ...
]
To use a tool, please use the following format:
```
Thought:Think what you need to solve, do you need to use tools?
Action:the tool name, should be one of [["ImageDescription", ... (other tool names)]]
Action Input:the input to the action
```
The response after utilizing tools should using the following format:
```
Tool Response:the results after call the tool.
```
If you already know the answer, or you do not need to use tools,please using the following format to reply:
```
Thought:the thought process to get the final answer
Final Answer:final answer
```
Begin!
<image>

[user] Where can someone buy the soda shown in this image?

[assistant] Thought:To answer the question, we first need to use the ImageDescription tool to identify the soda brand and flavor in the image, which is important for searching its availability.
Action:ImageDescription
Action Input:
```
{"image": <image>}
```

[user] Tool Response:The image shows a can of Postobon apple-flavored soda ("Manzana/Postobon") placed on a light blue surface, possibly a metal table. In the background, there is a colorful tiled mural featuring vibrant, geometric floral designs and fruits. The can is pink and red with the brand name prominently displayed in the center. The mural adds a lively and artistic vibe to the setting.

[assistant] Thought:... 

[user] Tool Response:...

... (action-observation loops)
\end{lstlisting}

\section{Demostrations}
\label{sec:suppdemo}

\begin{figure*}
    \centering
    \includegraphics[width=0.85\linewidth]{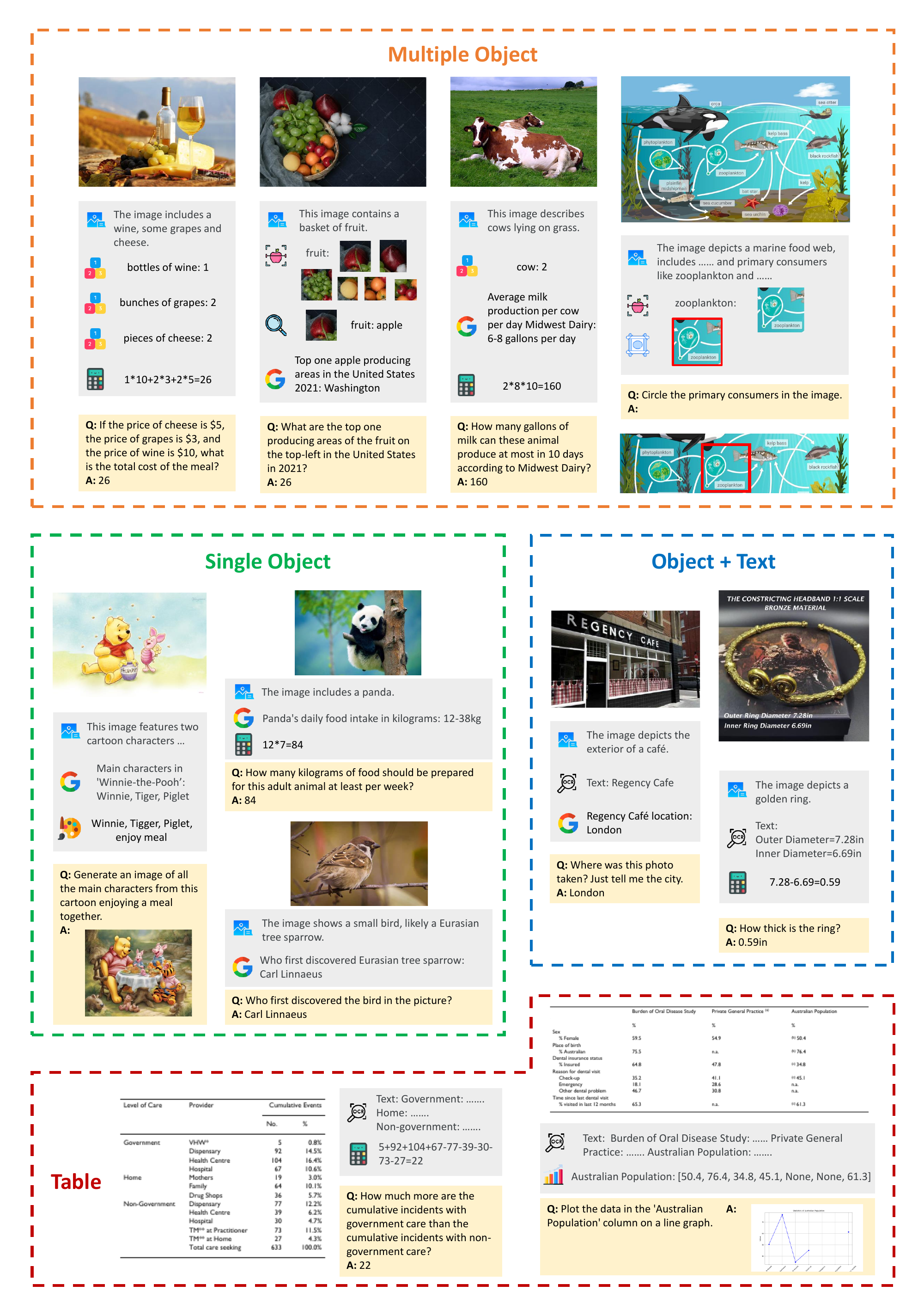}
    \caption{Demonstrations of real-world examples.}
    \label{fig:humanexample}
\end{figure*}

\begin{figure*}
    \centering
    \includegraphics[width=1.0\linewidth]{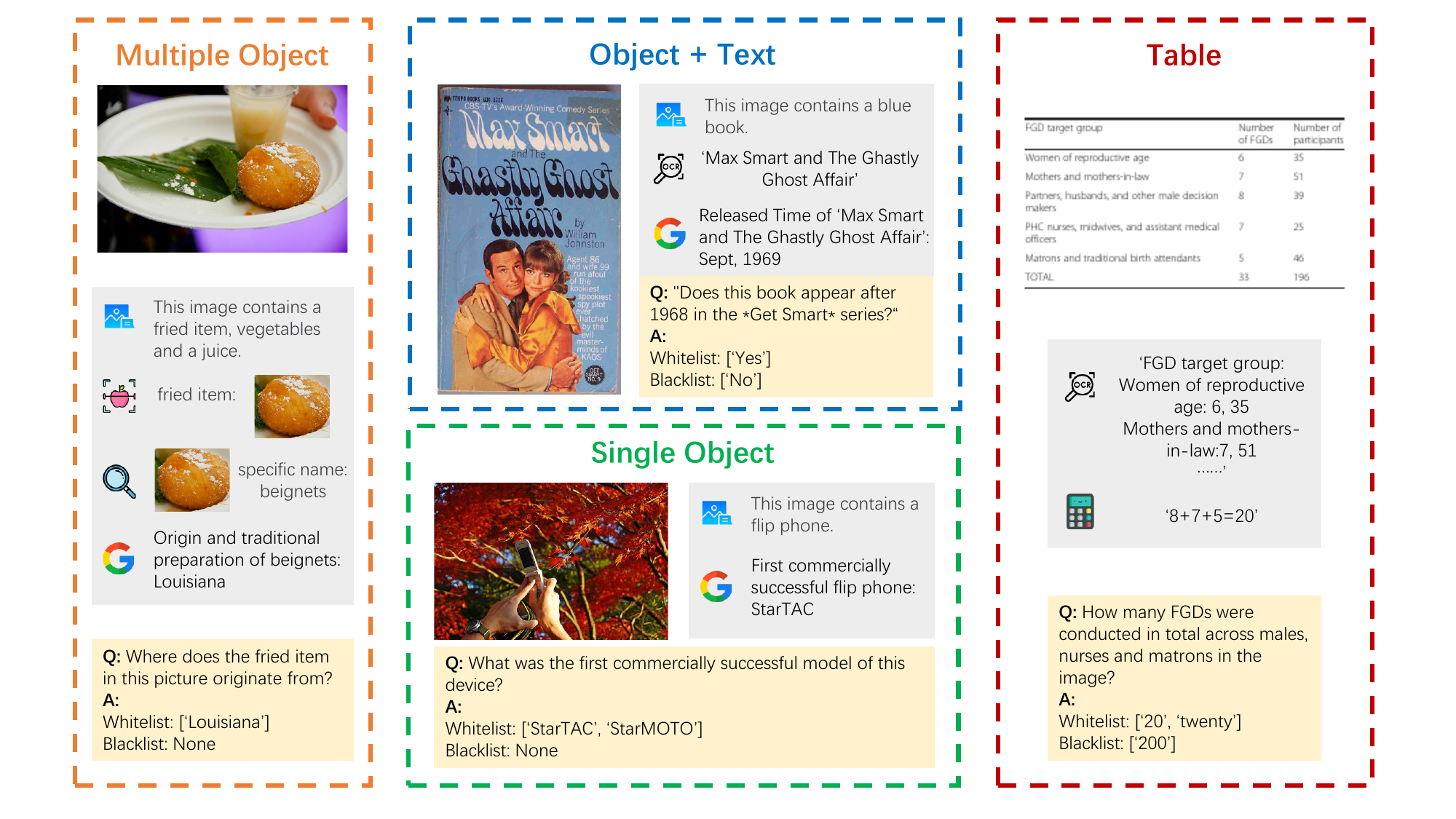}
    \caption{Demostrations of ToolVQA test set.}
    \label{fig:testset}
\end{figure*}

\cref{fig:humanexample} shows some of the real-world examples we used in the ToolEngine pipeline. \cref{fig:testset} demonstrates the data quality of the ToolVQA's test set. These data satisfy our definition of real scenarios and queries, and each sample requires more than one step of reasoning to solve. %

\end{document}